\documentclass[twoside]{article}
\usepackage[accepted]{aistats2016}
\usepackage{amsmath}
\usepackage{amsfonts}
\usepackage{amssymb}
\usepackage{amsthm}
\usepackage{color}
\usepackage{enumerate}
\usepackage{url}
\usepackage{enumitem}
\usepackage{graphicx}
\usepackage[numbers]{natbib} 
\usepackage{epsfig}
\usepackage[lofdepth,lotdepth]{subfig}

\usepackage{adjustbox}
\usepackage{float}

\usepackage{algorithm}
\usepackage{algpseudocode}

\usepackage{xcolor}
\usepackage{hyperref}
\hypersetup{
    colorlinks,
    linkcolor={red!60!black},
    citecolor={blue!75!black},
    urlcolor={blue!80!red}
}

\newcommand{\bp}[1]{{\boldsymbol \phi}(\mathbf{z}_{#1})}
\newcommand{\bphi}{{\boldsymbol \phi}}

\newcommand{\bR}{\mathbb{R}}
\newcommand{\bS}{\mathbb{S}}
\newcommand{\cZ}{\mathcal{Z}}
\newcommand{\bZ}{\mathbf{Z}}
\newcommand{\bz}{\mathbf{z}}
\newcommand{\tbz}{\ddot{\mathbf{z}}}
\newcommand{\bU}{\mathbf{U}}
\newcommand{\bu}{\mathbf{u}}
\newcommand{\bV}{\mathbf{V}}
\newcommand{\bv}{\mathbf{v}}

\newcommand{\ba}{\mathbf{a}}
\newcommand{\tba}{\tilde{\mathbf{a}}}
\newcommand{\bP}{\mathbb{P}}
\newcommand{\bx}{\mathbf{x}}

\newcommand{\by}{\mathbf{y}}

\newcommand{\cB}{\mathcal{B}}
\newcommand{\cA}{\mathcal{A}}

\setlist[itemize]{leftmargin=0.8cm}
\setlist[enumerate]{leftmargin=0.8cm}

\usepackage{mathtools}
\DeclarePairedDelimiter\ceil{\lceil}{\rceil}
\DeclarePairedDelimiter\floor{\lfloor}{\rfloor}

\newtheorem{lemma}{Lemma}

\let\oldref\ref
\renewcommand{\ref}[1]{(\oldref{#1})}

\begin{document}

%

%

\twocolumn[

\aistatstitle{Geometry Aware Mappings for High Dimensional Sparse Factors}\vspace*{-0.6cm}


\runningauthor{Bhowmik et al}



\aistatsauthor{Avradeep Bhowmik \And Nathan Liu}\vspace*{-0.1cm}

\aistatsaddress{ University of Texas at Austin, Austin, TX \And Google,  Mountain View, CA}\vspace*{-0.5cm}

\aistatsauthor{Erheng Zhong \And Badri Narayan Bhaskar \And Suju Rajan}\vspace*{-0.1cm}

\aistatsaddress{Yahoo! Labs, Sunnyvale, CA \And Yahoo! Labs, Sunnyvale, CA \And Yahoo! Labs, Sunnyvale, CA}\vspace*{-0.3cm}
]

\begin{abstract}
While matrix factorisation models are ubiquitous in large scale recommendation and search, real time application of such models requires inner product computations over an intractably large set of item factors. 
In this manuscript we present a novel framework that uses the inverted index representation to exploit structural properties of sparse vectors to significantly reduce the run time computational cost of factorisation models. We develop techniques that use geometry aware permutation maps on a tessellated unit sphere to obtain high dimensional sparse embeddings for latent factors with sparsity patterns related to angular closeness of the original latent factors. We also design several efficient and deterministic realisations within this framework and demonstrate with experiments that our techniques lead to faster run time operation with minimal loss of accuracy.
\end{abstract}

\section{INTRODUCTION}

Latent factor models like matrix 
\cite{fact1,fact2} and tensor \cite{tensor1,tensor2} factorisation are a ubiquitous class of techniques with a wide range of applications, mainly in personalised search and recommendation systems, where each user $i$ and each item $j$ is assumed to be associated with latent factors $\bu_i, \bv_j \in \bR^k$ respectively and the matrix of interactions (ratings, click/no-click, etc.) $\mathbf{R} = [r_{ij}]$ for the $i^{th}$ user with the $j^{th}$ item is modelled as the product of their respective  latent factors as $r_{ij} \sim \bu_i^\top \bv_j$ or a monotonic function thereof.

While substantial amount of work \cite{fact1, fact2, dist} has been dedicated to learning the latent factors given interaction data in a scalable manner, an often overlooked problem is the computational efficiency of deploying the learned factors for real time recommendation. 

The commonly used brute force retrieval of top-$\kappa$ relevant items for any user $i$ requires score computation of the corresponding latent factor $\bu_i$ with every single item factor $\bv_j \ \forall\ j = \{1, 2, \cdots N\}$, which is often an intractably large set. 
Pre-computing scores during the learning step is often impractical, for instance, in online news recommendation, where user interests change very rapidly and new items keep cropping up all the time. Moreover, while changing latent factors can be learned dynamically, arbitrary changes in latent factors would require updates to the entire set of pre-computed scores. 

A greatly preferable alternative would be to design a technique that automatically discards irrelevant items per user, and thereby significantly reduces the search space for top-$\kappa$ recommendations. This manuscript does exactly this, by exploiting structural properties of sparse vectors using the inverted index representation.\vspace*{-0.3cm}


\subsection{Sparse Factors and the Inverted Index Representation}
Suppose the factors for users and items were very sparse, inner product between factors with non-overlapping sparsity patterns (non-zero's in different indices) would compute to 0. In such a case for every user, relevant items would be such that the corresponding user factor and items factor have more or less matching sparsity pattern. 

The inverted index representation \cite{index1,index2}, widely used in information retrieval tasks, is particularly appropriate to exploit this property. In our setup, this involves storing the list of items using a data structure where each index is associated with all the items whose corresponding latent factors are non-zero in that index. 

During recommendation, for each user, we extract the set of indices $\mathcal{I}_\bu$ in which the corresponding user factor $\bu$ is non-zero, and retrieve using the inverted index notation, the set of items which are also non-zero in the corresponding indices in $\mathcal{I}_\bu$. Inner product computation is then required only over this significantly smaller set, rather than the full item set.


\subsection{Conflicting Sparsity Pattern}

Clearly, the success of the inverted index representation relies on using factors with significant ``conflict" in their sparsity patterns\footnote{Two sparse vectors have a conflicting sparsity pattern if the set of indices of non-zero elements for the two vectors are disjoint or have a very small intersection. For example $[9, 0, 8, 0, 0]$ and $[0, 6, 0, 7, 3]$ have non-zero elements in non-overlapping sets of indices $\{0, 2\}$ and $\{1, 3, 4\}$ respectively}.

Low dimensional factors are almost always dense, and to avoid losing too much information, introducing sparsity would need to be accompanied with an increase in dimensionality. 
Unfortunately, most learning algorithms that promote sparsity (like LASSO) cannot necessarily ensure conflicting sparsity patterns. A more reliable alternative is post-processing factors (possibly dense, and learned using any appropriate algorithm) to obtain high dimensional sparse embeddings such that for original factors that are ``close" to each other (high inner product), the corresponding sparse maps would have significant overlap in sparsity patterns, and vice versa.

To our knowledge, we are the first to explicitly tackle this exact problem setup, to use the inverted index representation to discard irrelevant factors with conflicting sparsity pattern. Our main contributions are as follows:
\begin{enumerate}
\item We introduce a novel framework consisting of a meta-algorithm that uses geometry aware permutation maps on a tessellated unit sphere to obtain sparse embeddings for latent factors with sparsity patterns related to angular closeness of the original factors. 
\item We provide several deterministic realisations for the meta-algorithm that are efficient with respect to time and space complexity and satisfy desirable properties
\item We demonstrate the efficacy of our methods with extensive experimental evaluation \vspace*{-0.3cm}
\end{enumerate}

\subsection{Notation}
$\bS^k$ refers to the surface of the $k$-dimensional Euclidean unit sphere. We use (usually subscripted) blackboard bold font $\bP$ to denote permutations, always in $p$-dimensional space unless indicated otherwise, where $p>k$. 
Finally, we use $\bphi : \bS^k \mapsto \mathbb{R}^p$ to refer to our sparse mapping function that maps factors $\bz$ on the $k$-dimensional hypersphere to a sparse $p$-dimensional vector $\bphi(\bz)$. Zero padded vectors are denoted by diacritics as $\tbz$.


\section{PROBLEM SETUP}

Consider the following setup. We are given a set of $N$ factors $[\bz_1; \cdots ;\bz_N] = \bZ \in \cZ \subseteq \bS^k$, where $\bS^k = \{\mathbf{x} \in \bR^k: \|\mathbf{x}\|_2 = 1\}$ is the unit sphere in $k$-dimensional Euclidean space $\bR^k$. For example, in a recommendation setting the set of factors $\bZ$ could be the concatenated set $\bZ = [\bU; \bV]$ of user features $\bU$ and object features $\bV$.

The ``compatibility" between two factors $\bz_i, \bz_j$ is measured as $r_{ij} = \bz_i^\top \bz_j$. In the context of recommendations, $\bz_i = \bu_i$ is the $i^{th}$ user factor, and $\bz_j = \bv_j$ is the $j^{th}$ item factor, and $r_{ij}$ is the interaction (rating, click/no-click).

This notion of ``compatibility" between factors on the unit sphere $\bS^k$ is captured by the angular distance\footnote{alternatively, one minus standard cosine similarity} metric $d(\cdot, \cdot)$ which is defined for any two factors $\bx, \by \in \bR^k$ as $$d(\bx, \by) =  1 - \frac{\bx^\top \by}{\|\bx\|_2\|\by\|_2}$$ 

For factors $\bz_i, \bz_j \in \bZ$, $\|\bz\|_2 = 1$, therefore $d(\bz_i, \bz_j) = 1 - \bz_i^\top \bz_j = 1 - r_{ij}$. Clearly, factors which are more compatible have a low angular distance in their Euclidean vector representations and vice versa. 

The objective is to find a map $\bphi: \cZ \mapsto \bR^p$ that maps factors in $\cZ$ to sparse vectors in a $p$-dimensional space $\bR^p$, where $p > k$. 

As described earlier, the inverted index representation is useful in extracting vectors which have overlapping sparsity patterns. Hence, the mapping $\bphi$ should be such that if two factors $\bz_i$ and $\bz_j$ have a low angular distance between them, their corresponding mappings $\bphi(\bz_i)$ and $\bp{j}$ should have similar sparsity patterns. Conversely, if $\bz_i$ and $\bz_j$ have a high angular distance between them, their corresponding mappings $\bp{i}$ and $\bp{j}$ should have conflicting sparsity patterns. 

\section{A GEOMETRY AWARE SCHEMA FOR SPARSE MAPPING}

\begin{figure*}[!htb]
\centering
\subfloat[subfig1][Tessellating unit sphere into regions, and marking each $\bz$]{
\includegraphics[width=0.26\textwidth, height = 4.5cm]{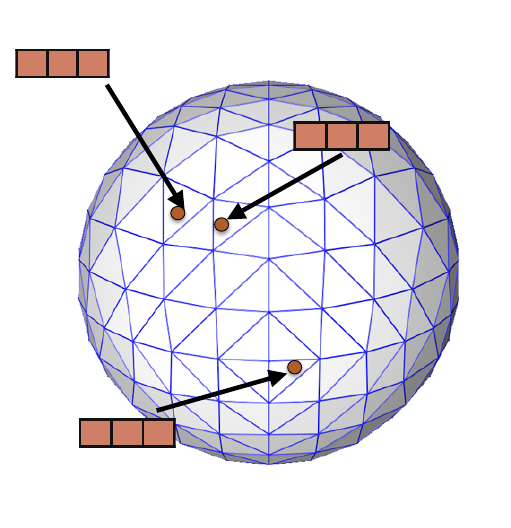}
\label{fig:tess}}\hfill
\subfloat[subfig2][Padding with zeros, to obtain $\ddot{\bz} \sim \{\bz ; \mathbf{0}\}$ ]{
\includegraphics[width=0.26\textwidth, height = 4.5cm]{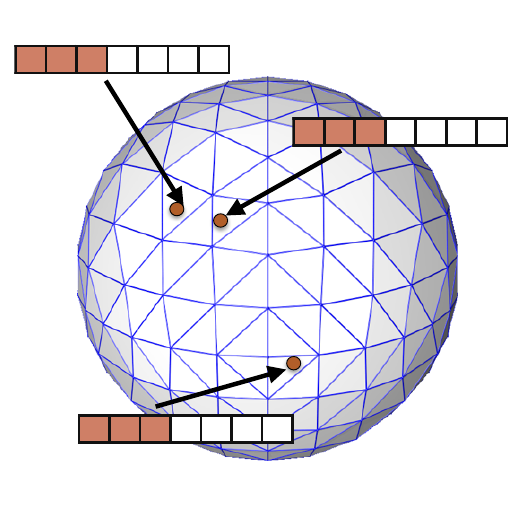}
\label{fig:pad}}\hfill
\subfloat[subfig3][Applying region based permutation, to get $\bphi(\bz) = \bP(\ddot{\bz})$]{
\includegraphics[width=0.26\textwidth, height = 4.5cm]{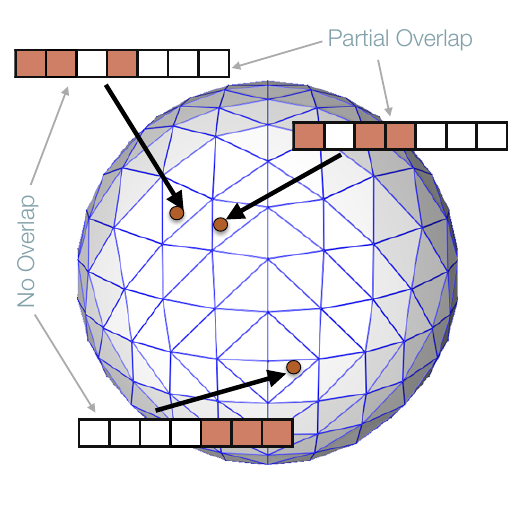}
\label{fig:perm}}
\caption{A pictorial representation of the sparse mapping technique- (a) Tessellating the unit sphere and associating each factor with its corresponding tile (b) Padding each factor with zeros to make it $p$-dimensional (c) Applying to each zero-padded vector the permutation specific to its tile on the tessellated unit sphere}
\label{fig:Pict}
\vspace*{-0.5cm}
\end{figure*}

The requirements on $\bphi(\cdot)$ elaborated on in the preceding section can be captured effectively with the following intuitive observation. Suppose there exists a mapping between sparsity patterns and regions on the surface of the unit sphere such that neighbouring regions of the unit sphere get similar sparsity patterns and vice versa. Then the requirements on the mapping function can be satisfied by setting $\bphi$ to map every factor with the sparsity pattern that depends on which region on the unit sphere the factor lies.

Our proposed framework uses this intuition to design a mapping function $\bphi(\cdot)$ that maps compatible (angularly close) factors to overlapping sparsity patterns and incompatible factors to conflicting sparsity patterns. The steps involved are described below (and summarised as the meta-algorithm in Algorithm \ref{alg:meta}). We show concrete realisations for the meta-algorithm in section \ref{sec:examples}.\vspace*{-0.3cm}

\subsection{Defining a Schema}

First we define a schema consisting of tessellating the unit sphere and a permutation map for each tile or region of the tessellated unit sphere.

\textbf{Step I: Tessellating the unit sphere}: 

A natural way of defining regions corresponding to angular distance is via tiles on the surface of a tessellated unit sphere. An $M$-order tessellation for our purpose is specified completely by a set of $M$ distinct tessellating vectors $\Gamma = \{\ba_i : i = 1, 2, \cdots M, \|\ba_i\| = 1\} \subset \bS^k$. Each tile or region $\gamma_\ba$ associated with a specific tessellating vector $\ba$ under this schema is defined simply as the set of points which are closest to the said tessellating vector, that is, $$\gamma_\ba = \{\bx \in \bS^k: d(\gamma_\ba, \bx) \leq d(\gamma_{\ba'}, \bx) \ \forall \ \ba' \in \Gamma, \ba \neq \ba'\}$$

Therefore, the boundary between regions under such a tessellating scheme consists of points on the unit sphere which are equidistant from two or more tessellating vectors. Note that this is similar to the concept of a Voronoi tessellation in a metric space.

\textbf{Step II: Associating every region with a permutation}:

The next step is to associate a permutation $\bP_\ba$ with each $\ba \in \Gamma$, we denote the set of all such permutations by $\bP_\Gamma = \{\bP_\ba : \ba \in \Gamma\}$. The main requirement for this mapping is that nearby tessellating vectors should get mapped to similar permutations. An informal existence argument can be made for this using the fact that the set of permutations, when represented as the vertices of a Birkhoff polytope \cite{Birkhoff}, can be embedded on the surface of a unit hypersphere \cite{direction}. 


\subsection{Processing Factors based on the Schema}

Given a schema $(\Gamma, \bP_\Gamma)$, defining the map $\bphi(\cdot)$ for a set of factors $\bZ$ consists of the following steps

\textbf{Step I: Associating every factor to a region}:

The associated region for a factor $\bz$ is specified by the closest (in angular distance) tessellating vector $\ba_\bz \in \Gamma$ to the factor, as determined by the following optimisation problem 
\begin{equation}\label{eq:project}
\ba_\bz = \text{arg}\min_{\ba \in \Gamma} d(\ba, \bz)
\end{equation}

This is, in general, a difficult optimisation problem, with a search space over an intractably large discrete domain, but as we shall see in the succeeding sections, many tessellating schemata admit efficient solutions, exact or approximate, to this problem.

\textbf{Step II: Zero-padding factors}:

The next step is simply to zero-pad the $k$-dimensional vector with $p-k$ zeros to make it $p$-dimensional. Denote the zero-padded factor for $\bz$ by $\tbz$.

\textbf{Step III: Applying region specific permutation}

Finally, we apply region specific permutations as defined by $\bP_\Gamma$ to the zero-padded vector. Say $\ba_\bz$ is the tessellating region associated with a factor $\bz$, and $\bP_{\ba_\bz}$ is the corresponding $p$-dimensional permutation associated with $\ba_\bz$, then the mapping $\bphi$ maps $\bz$ to $\bR^p$ by using the permutation operator $\bP_{\ba_{\bz}}$ on the zero-padded factor $\tbz$ as 
\begin{equation}\label{eq:applyperm}
\bphi(\bz) = \bP_{\ba_{\bz}}(\tbz)
\end{equation}

A pictorial\footnote{parts of the image adapted from the web} depiction of the technique has been shown in Figure \ref{fig:Pict}. With proper selection of tessellation schema and permutation map, factors in nearby tiles have similar sparsity patterns (high overlap in non-zero entries) in their sparse maps and vice versa.

\begin{algorithm}
\caption{Sparse-mapping meta-algorithm}\label{alg:meta}
\begin{algorithmic}[1]
\Procedure{Schema}{M}

\State Define tessellating set of $M$ vectors $\Gamma \subset \bS^k$
\State Define permutation map $\mathbb{P}_{\ba}$ for each $\ba \in \Gamma$
\State \textbf{return} $\Gamma, \mathbb{P}_\Gamma$
\EndProcedure
\Statex
\Procedure{ProcessFactors}{$\Gamma, \mathbb{P}_\Gamma, \bZ$}
\For{each $\bz \in \bZ$}
\State associate region as $\ba_\bz = \text{arg}\min_{\ba \in \Gamma} d(\ba, \bz)$
\State zero-pad $\tilde{\bz} = [\bz; 0]$
\State apply permutation to get $\bphi(\bz) = \mathbb{P}_{\ba_\bz}(\tilde{\bz})$
\EndFor
\State \textbf{return} $\bphi(Z)$
\EndProcedure
\end{algorithmic}
\end{algorithm}


\subsection{Desiderata for a good schema}

Apart from finding effective tessellation schemata and permutation maps for each tessellation schema, there are multiple challenges that need to be taken into consideration when designing a particular instance for this meta algorithm. Note that for a given schema, processing the factors involves two potentially computationally intensive steps defined in equations \ref{eq:project} and \ref{eq:applyperm}. Hence, any schema for tessellation and permutation should be such that both \ref{eq:project} and \ref{eq:applyperm} should be efficiently computable. Another concern would be controlling the storage complexity.

The first natural technique that is immediate as a tessellation schema is hypersphere point picking where random points are picked uniformly distributed on the surface of the unit hypersphere (see for instance \cite{sphere1,sphere2,sphere21,sphere3}). Owing to \cite{spheregauss}, we have a simple method of doing this by generating $M$ independent and identically distributed points from the standard $k$-dimensional multivariate Gaussian distribution and normalising them. Since the standard Gaussian distribution is spherically symmetric, the resulting points are uniformly distributed on the surface of the unit hypersphere.

However, it is immediately apparent that for any randomised schema for $\Gamma$, solving the optimisation problem \ref{eq:project} for any $\bz \in \bZ$ 
would require an exhaustive search involving an explicit computation of distance scores with every $\ba \in \Gamma$. Since $M$ can be really large (often super exponential in $k$, as we shall see in section \ref{sec:examples}), this is computationally infeasible. Moreover, because this requires explicit generation and storage of all $M$ tessellating vectors, the space complexity required is also prohibitively high.

A similar argument holds for the design of the permutation map as well. The space of permutations over $p$ coordinates is $O(2^{plogp})$ in $p$, and assigning permutations one by one to each of the $M$ tessellating vectors would be infeasible.

A one-stop solution to both of these problems is to have a deterministic function-based schema for both the tessellation and permutation map. That is, given a factor $\bz \in \bZ \subset \bS^k$, finding the tessellating vector closest to it should involve evaluating a function of $\bz$ which is computationally efficient in terms of $k$, and for a given tessellating vector $\ba \in \Gamma$, computing the corresponding permutation $\bP_\ba$ should involve evaluating a function of $\ba$ which can be done efficiently in terms of $p$. Therefore, mapping for each factor can be done separately in a two-step process without additional explicit storage or computation of $\Gamma$ or $\bP_\Gamma$.


\section{DETERMINISTIC SCHEMATA}\label{sec:examples}\vspace*{-0.2cm}

In this section, we describe some concrete realisations for the general schema described earlier. In particular, we specify simple techniques for tessellating the unit sphere, as well as associating a permutation map with each of these tessellations, and generalise each of these techniques to define a broader class of realisations for our schemata. We will show that finding $\bphi(\bz)$ given $\bz$ using each of these techniques only involves two efficient deterministic function computations, hence they avoid all of the pitfalls described in the previous section. Note that obtaining $\bphi(\bz)$ for each $\bz$ can be done separately for each $\bz$ in parallel.\vspace*{-0.2cm}

\subsection{Tessellating the Unit Sphere}

The first step in the meta-algorithm is to tessellate the unit sphere. We start off by describing a simple scheme in section \ref{sec:tess} and then generalise the idea to specify a larger class of tessellation schemes in section \ref{sec:tessgen}.

\subsubsection{Directional tessellation}\label{sec:tess}

Consider the ternary base set $\cB = \{-1, 0, 1\}$. Say $\cA = \cB^k - \{0\}^k$ is the set of all\footnote{For example, for $k=2$, we have $\cA = \{[-1, -1], [-1, 0], [-1, 1], [0, 1], [1, 1]\}$.} non-zero $k$-length ternary vectors formed out of elements of $\{-1, 0, 1\}$. The tessellating vector set is formed from the normalised versions of these vectors, that is
\begin{equation}\label{eq:tess}
\Gamma = \{\ba = \frac{\tilde{\ba}}{\|\tilde{\ba}\|}: \tilde{\ba} \in \cA\}
\end{equation}  
Clearly in this case, $M = |\Gamma| = 3^k - 1$. However, finding the closest tessellating vector given any $\bz \in \bZ$ can be done efficiently.\footnote{Note that the na\"{i}ve algorithm that thresholds every index of $\bz$ on -0.5 or 0.5 to get an element of $\cB$ does not give an exact solution since we are working with angular distance metrics}

\begin{lemma}\label{lem:tess} 
Given a factor $\bz \in \bZ \subset \bS^k$, solving equation \ref{eq:project} for $\Gamma$ as in \ref{eq:tess} can be performed efficiently in $O(k logk)$ time using Algorithm \ref{alg:tess}, and requires no explicit storage of the tessellating set $\Gamma$.
\end{lemma}

The time complexity of Algorithm \ref{alg:tess} is $O(klogk)$ since the limiting operation is a sorting operation which can be done in $O(klogk)$ time. Proof of correctness of this algorithm is presented in the supplementary material.

\begin{algorithm}
\caption{Region Specification on $\Gamma$ defined on $\cB$}\label{alg:tess}
\begin{algorithmic}[1]
\Procedure{TessVector}{$\bz$}
\State Sort $\bz$ in desc. order of abs. value to get $\bz_\downarrow$
\State Let $\pi$ be the sorting order, that is, $$\bz_\downarrow^j = |\bz^{\pi(j)}| \geq \bz^{\pi(j+1)}| = \bz_\downarrow^{j+1} \ \forall \ j$$
\State Compute scaled cumsum $\bz_s$ from $\bz_\downarrow$ as
\For{$\iota = 1, 2, \cdots k$}
\State Set $\bz_s^\iota = \frac{\sum_{j = 1}^\iota \bz_\downarrow^j }{\sqrt{\iota}}$
\EndFor
\State Let $\iota^* = \text{argmax}_\iota \bz_s^\iota$ be the index of the maximum value of $\bz_s$
\State Define the index set (where $\bz_\downarrow^{\pi^{-1}(k)} = |\bz^{k}|\ \forall k$) $$I_\bz = \{\pi^{-1}(1), \pi^{-1}(2), \cdots \pi^{-1}(\iota^*)\}$$
\State Compute the tessellating vector $\ba_\bz$ as
\begin{equation*}
\ba_\bz^\iota = 
\begin{cases}
\frac{sign(\bz^\iota)}{\sqrt{|I_\bz|}} \ \ \text{ if } \iota \in I_\bz\\
0 \ \ \text{ otherwise }
\end{cases}
\end{equation*}
\State \textbf{return} $\ba_\bz$
\EndProcedure
\end{algorithmic}
\end{algorithm}

\subsubsection{Directional tessellation using D-ary Base set}\label{sec:tessgen}

This is exactly the same as the previous case, except the tessellation is done with a $D$-ary base set $\cB_D = \{-1, -\frac{D-1}{D}, \cdots , -\frac{1}{D}, 0, \frac{1}{D}, \cdots , \frac{D-1}{D}, 1\}$ instead of a ternary base set- the corresponding vector set is $\cA_D = \cB_D^k - \{0\}^k$ and tessellating set is $\Gamma_D = \{\ba = \frac{\tilde{\ba}}{\|\tilde{\ba}\|}: \tilde{\ba} \in \cA_D\}$. Clearly, the ternary base set $\{-1, 0, 1\}$ is the same as $\cB_D$ with $D = 2$.

The algorithm \ref{alg:tess} for ternary base sets no longer applies directly to the $D$-ary case. In fact, getting an exact solution is, in general, difficult for this schema. However, we can still get an $\epsilon$-approximation to the closest tessellating vector (the corresponding algorithm is provided in the supplement).

\begin{lemma}\label{lem:tessD}
For any vector $\bz \in \bZ$, say $\ba_\bz^*$ is the true solution to equation \ref{eq:project} with $\Gamma$ obtained the $D$-ary base set. Then, a tessellating vector $\tilde{\ba}_\bz$ can be obtained such that $d(\ba_\bz^*, \tilde{\ba}_\bz) \leq \epsilon$, where $\epsilon \sim O(k/D^2)$, using an algorithm that takes $O(k)$ time, and requires no explicit storage of the tessellating set $\Gamma_D$.
\end{lemma}

Therefore, if $D \gg \sqrt{k}$, the true tessellating vector can be obtained within a very small tolerance. The algorithm to obtain this and the proof of this lemma is provided in the supplementary material. Clearly, this schema generates a finer tessellation of the unit sphere with increasing $D$.\vspace*{-0.3cm}

\subsection{Sparse Mapping in High Dimensions}

The next step in the method is to assign a sparsity pattern to every tessellating vector. We start off by describing a simple permutation map in section \ref{sec:perm} and then generalise the idea to define a larger class of maps in section \ref{sec:permgen}. Note that each sparsity pattern can also be defined as a function (that depends on the corresponding tessellating vector $\ba_\bz$) that maps a coordinate of $\bz$ to a specific coordinate of $\bphi(\bz)$.

\subsubsection{One Hot Encoding}\label{sec:perm}
A simple encoding scheme is the following. Consider the ternary tessellation scheme as in Section \ref{sec:tess}. Note that for tessellating vectors obtained from a ternary scheme, there is a one-one correspondence between each $\tilde{\ba} \in \cA$ and each $\ba \in \Gamma$ since $$\ba^j = \frac{sign(\tilde{\ba}^j)}{(\text{num of non-zeros in } \tilde{\ba})^{1/2}}\ \ \forall \ j$$

Therefore, a simple permutation scheme for this tessellation can be obtained for $p = 3k$ as follows. For every $t = 1, 2, \cdots k$ and $i = 1, 2, \cdots p$ set
\begin{equation*}
\bphi(\bz)^i = [\bP_{\ba_z}(\tbz)]^i =
\begin{cases}
\bz^t \ \ \text{ if } i = 3t \text{ and } \tilde{\ba}_\bz^t = 1 \\
\bz^t \ \ \text{ if } i = 3t + 1 \text{ and } \tilde{\ba}_\bz^t = 0 \\
\bz^t \ \ \text{ if } i = 3t + 2 \text{ and } \tilde{\ba}_\bz^t = -1 \\
0 \ \ \text{ otherwise }\vspace*{-0.3cm}
\end{cases}
\end{equation*}

This basically pads each coordinate of $\bz$ with two extra zeros and permutes it within each 3-index segment thus obtained depending on the value of the corresponding coordinate of $\tilde{\ba}_\bz$. A similar schema can be obtained for $p = Dk$ for the $D$-ary tessellation.

The permutation thus obtained is related to the geometry of the tessellation- for any two $\ba_1, \ba_2 \in \Gamma$, the Kendall-Tau distance\footnote{minimum number of pairwise inversions required to convert one permutation to another, see \cite{KDT}} between their corresponding permutations $\bP_{\ba_1}, \bP_{\ba_2}$ is exactly equal to the $\ell_1$ distance between the unnormalised vectors $\tilde{\ba}_1, \tilde{\ba}_2$. A bound for Spearman's footrule can also be obtained\cite{KDT}. 

It is easy to see that this mapping also has the desirable property that for any $\bz \sim \ba$ and $\bz' \sim \ba'$, suppose for a particular index $j \in \{1, \cdots k\}$, we have that $\tau_j$ and $\tau_j'$ are the corresponding index map for $\bphi(\cdot)$. That is, $\bphi(\bz)^{\tau_j} = \bz^j$ and $\bphi(\bz')^{\tau_j'} = {\bz'}^{j}$. Then, $\tau_j = \tau_j'$ if and only if $\ba_j = \ba_j'$. Moreover, the list of possible $\tau_j$ is unique for any $j$ and depends only on $j$, and not on $\ba$. This ensures that sparsity patterns overlap only for neighbouring tessellating regions, uniformly.\vspace*{-0.2cm}

\subsubsection{Parse Tree Based Encoding}\label{sec:permgen}

A more general scheme for this can be obtained in the following manner. Consider the ternary tessellation scheme of Section \ref{sec:tess}. Start with a $\bphi(\bz)$ as a $p$-length vector of all zeros. For constructing a parse-tree of depth $\delta$, initialise by mapping the first $\delta-1$ coordinates of $\bz$ to specific coordinates of $\bphi(\bz)$ using, say, the one-hot encoding scheme. At each subsequent step $j = \delta, \delta+1, \cdots k$, use a sliding window of size $\delta$ to read the unnormalised tessellating vector $\tba$, such that $\delta$ coordinates are read at a time. At step $j$ of the reading process, we read the segment $\tba_\delta^j = [\tba^{j - \delta}, \cdots , \tba^j]$. 

Since each coordinate of $\tba$ can take three possible values $\{-1, 0, 1\}$, we can construct a parse tree of depth $\delta$ containing $3^\delta$ leaf nodes which is traversed based on the $\delta$-length segment thus read. Thus, each non-leaf node at depth $t \in \{0, 1, \cdots (\delta-1)\}$ branches out into three child sub-trees corresponding to $\tba^{j-\delta + t}$ being -1, 0 or 1. 

Map the remaining coordinates of $\bphi(\bz)$ in the following way. At each step $j$ of the reading process, maintain an index counter $\tau_j \in \{1, 2, \cdots p\}$ that defines the current index at time $j$. Associate each leaf node of the parse tree with a corresponding ``action" $f(\cdot)$ to perform on the counter to move it to the current index. That is, given the previous index $\tau_{j-1}$ and the segment $\tba_\delta^j = [\tba^{j - \delta}, \cdots , \tba^j]$ read at time step $j$, compute the next position for the counter as $\tau_j = f(\tau_{j-1}; \tba_\delta^j)$. Finally, map the $j^{th}$ coordinate of $\bz$ to the current index of $\bphi(\bz)$ as $\bphi(\bz)^{\tau_j} = \bz^j$. Repeat the same process for each step $j = \delta, \delta+1, \cdots k$.

For example, the one-hot encoding is a special case of this with $\delta = 1$. Suppose we have already mapped $\bz^1$ to $\bz^{j-1}$. At step $j$, if the leaf node corresponding to $\ba_\bz^j$ is 1, the corresponding action is to set the counter $\tau_j$ at time $j$ to the coordinate $\tau_j = 3j$. Similarly, if $\ba_\bz^j$ is a 0, we set the counter to the coordinate $\tau_j = 3j + 1$ and if it is -1, we set $\tau_j = 3j+2$. The corresponding coordinate of $\bphi(\bz)$ gets mapped as $\bphi(\bz)^{\tau_j} = \bz^j$.

Other more complicated actions $f(\cdot)$ are possible. In particular, we describe some examples in the supplement which have the property that for any $\ba, \ba'$ at any step $\tau_j = \tau_j'$ if and only if $[\ba^{j-t}, \cdots \ba^j] = [{\ba'}^{j-t}, \cdots {\ba'}^j]$ for some $t > \delta$. Moreover, list of possible $\tau_j$ for any $j$ depends uniquely only on $j$, and not on $\ba$ itself. The parse-tree procedure can be extended to the case of a $D$-ary tessellation scheme in a simple manner by considering parse trees where each non-leaf node has $D$ child nodes.

The mapping procedure is deterministic and time efficient- suppose each action $f(\cdot)$ in an $O(1)$ operation (most actions like shifts, etc. are of this form), the effective time complexity is $O(k\delta)$. The final space complexity of storing $\bphi(\bz)$ for each $\bz$ is no more than $O(klogp)$ using the inverted index representation.\vspace*{-0.2cm}


\begin{figure*}[!htb]
\centering
\subfloat[subfig1][Histograms of percentage of discarded items over users]{
\includegraphics[width=0.5\textwidth, height = 6.0cm]{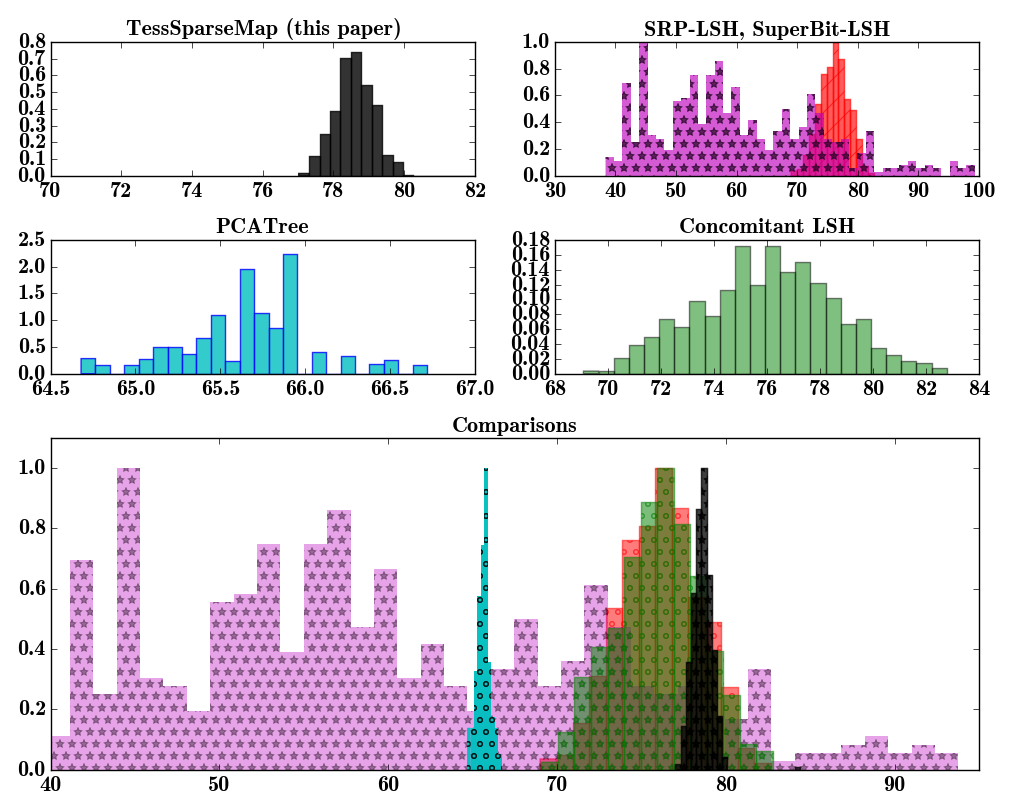}
\label{fig:Simspar}}
\subfloat[subfig2][Recovery Accuracy for Synthetic Data]{
\includegraphics[width=0.5\textwidth, height = 6.0cm]{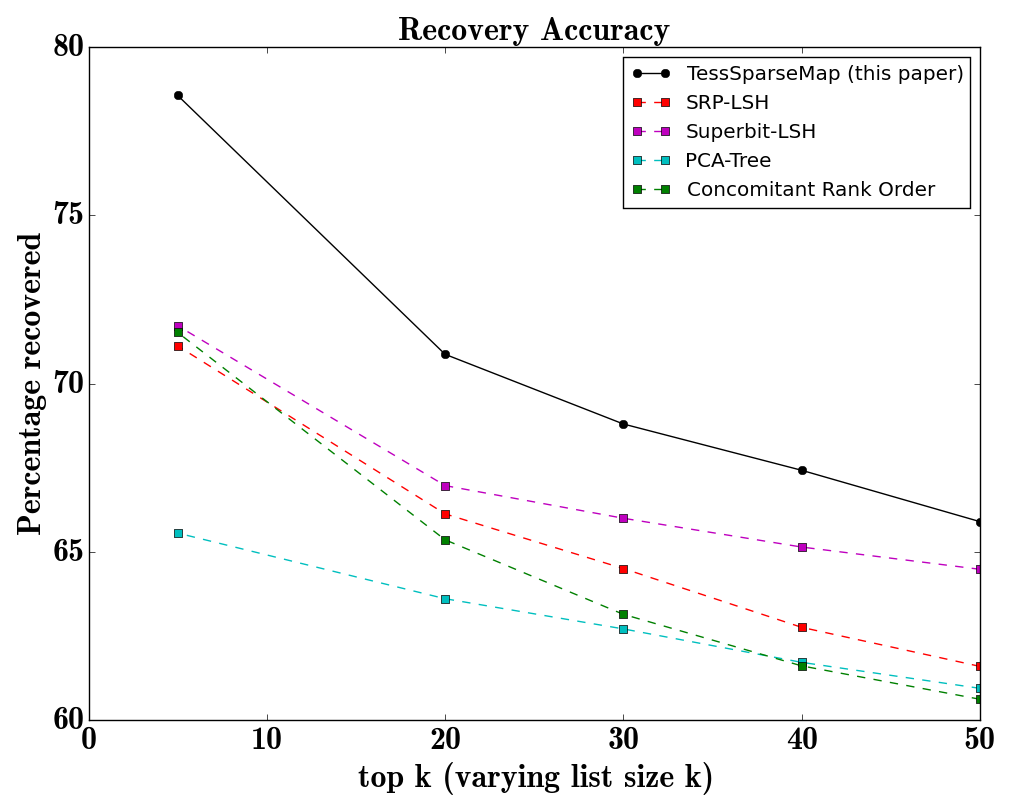}
\label{fig:SimAcc}}
\caption{Synthetic Data: our method discards more items on average with lower variance across users while having higher recovery accuracy (histogram y-axes scaled for uniformity)}
\label{fig:ResSim}
\vspace*{-0.5cm}
\end{figure*}

\section{DISCUSSION}\vspace*{-0.3cm}

We would like to note that while the preceding discussion assumed normalised factors, we do not actually need the factors (or, indeed, even the tessellating vectors) themselves to have unit norm, since we are anyway looking at the angular distance metric. In particular, Algorithm \ref{alg:tess} for finding the closest (in terms of angular distance) tessellating vector $\ba$ for a given factor $\bz$ is scale invariant in terms of $\bz$ as well as the set of $\ba$.

Next, note that our schema is generic and does not depend on the specific learning algorithm used for the latent factors. Moreover, because of the geometric nature of our framework, it can work for all kinds of factors irrespective of spherical symmetry properties of the factor distribution. For factors which are known to have clustered form, a simple extension of our algorithm would involve a non-uniform tessellation scheme with finer granularity near the cluster centres. \newpage A discussion on non-uniform tessellation is provided in the supplementary material.\vspace*{-0.3cm}


\subsection{Related Work}\vspace*{-0.2cm}

The closest line of work to our method is locality sensitive hashing \cite{LSH0,LSH1} which looks at approximate nearest neighbour extraction. However, while hashing is mostly concerned with dimensionality reduction (faster score computation by decreasing the effective $k$), our problem has more to do with direct reduction of effective search space (faster retrieval by reducing effective $N$); in fact, the two methods are independent enough that they can augment each other. Hash functions have been developed for different distance metrics like Hamming distance \cite{LSH1}, Euclidean distance \cite{LSH2}, Jaccard Similarity \cite{minhash}, etc. The most popular hash function for angular similarity is the sign-random-projection hash (SRP-LSH) \cite{SRP-LSH} which generates random hyperplanes and assigns to a factor the sign of the projection of that factor on each hyperplane. A more recent alternative to this called Superbit-LSH \cite{SuperLSH} orthogonalises the random vectors before projection. Another variation uses $l$ concomitant rank order statistics \cite{concom} instead of signed projection to compute an $l$-ary hash code. Yet another line of work computes hash functions by constructing spatial partitioning trees, specifically the PCA-tree \cite{PCAtree} which splits each factor at the median along principal eigenvectors.

For standard hashing methods, the usual way is to extract relevant items for each user by computing the Hamming distance with the hash functions of corresponding item factors and returning the closest items. In our setup that is not feasible since that would defeat the entire purpose of not having to compare against every item. These methods would apply to our setup by computing exact hash matches using tree-based data structures. However, each instance of LSH divides factors into regions with rigid boundaries, which tends to throw away too many items\footnote{in our experiments, LSH is boosted by coalescing all items collected by multiple instances of random hashing}, especially for factors at the edges. In contrast, since a geometry aware schema is tuned to each factor separately, our method by design also captures similarity with overlapping regions and soft boundaries.

Uniformly tessellating the unit sphere deterministically is a hard problem. Various heuristics \cite{tess0,tess1,tessicosa} exist that use, for example, arguments from physics to find minimum energy configurations for charges on a sphere. Embedding permutations on the unit sphere is even more difficult, see for instance \cite{direction}. However, unlike our methods, all of these techniques require computationally expensive exhaustive search approaches for finding the correct tessellation for a given factor, or finding the appropriate permutation map.\vspace*{-0.2cm}



\begin{figure*}[!htb]
\centering
\subfloat[subfig1][Histograms of percentage of discarded items over users]{
\includegraphics[width=0.5\textwidth, height = 6.0cm]{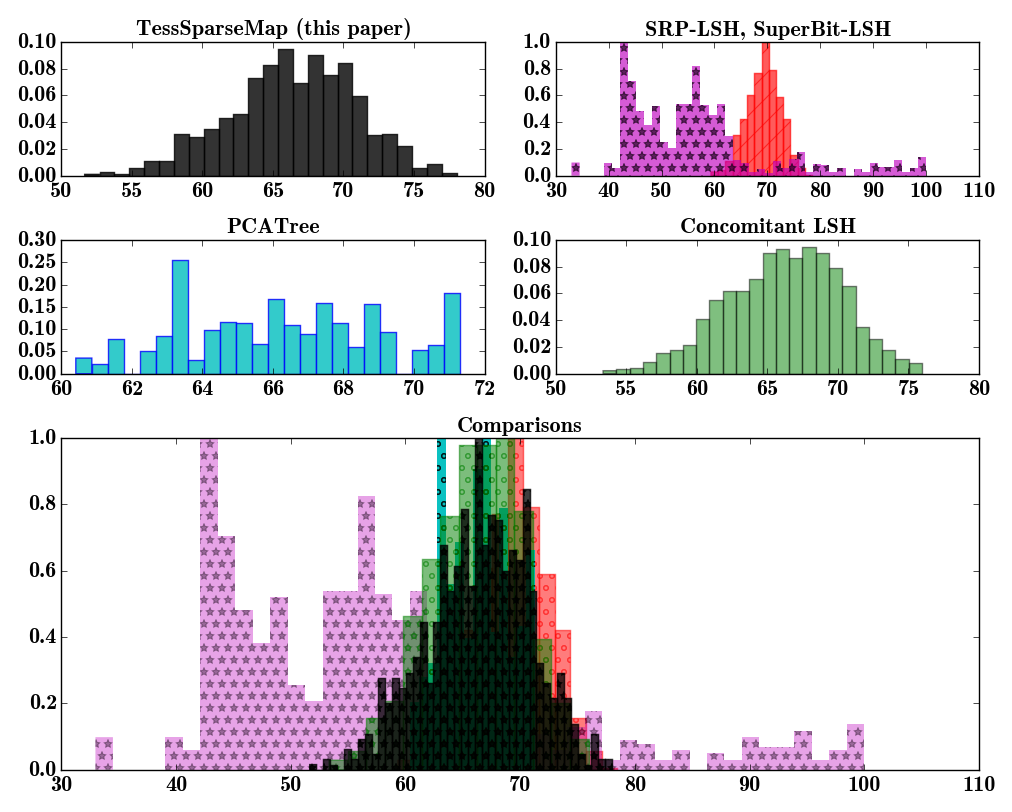}
\label{fig:MLspar}}
\subfloat[subfig2][Recovery Accuracy for MovieLens Data]{
\includegraphics[width=0.5\textwidth, height = 6.0cm]{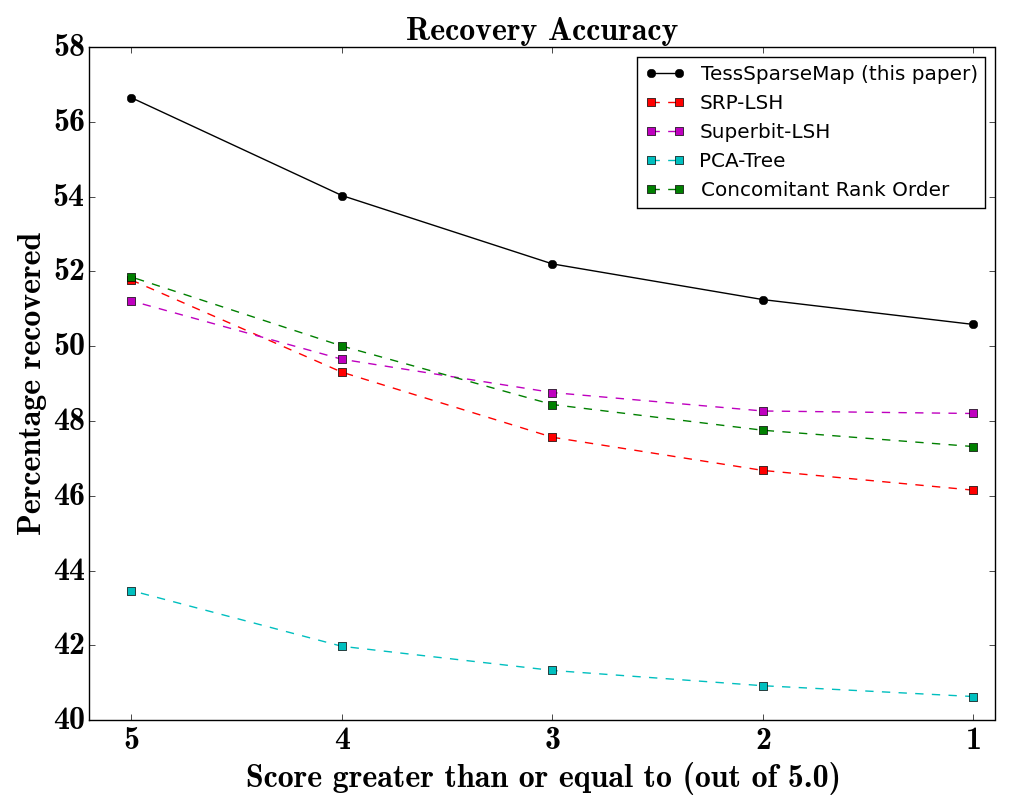}
\label{fig:MLAcc}}
\caption{MovieLens Data: for comparable percentage of discarded items, recovery accuracy is much higher for our method compared to baselines (histogram y-axes scaled for uniformity)}
\label{fig:ResML}
\vspace*{-0.5cm}
\end{figure*}

\section{EXPERIMENTS}\vspace*{-0.3cm}

We perform experimental evaluation of our procedure on both synthetic data and the MovieLens100k dataset \cite{ML} consisting of ratings compiled from the MovieLens website. 
Using recommendation system terminology, one set of factors will be referred to as user factors, the other set as item factors and the inner product between user and item factors will be referred to as the rating.


For our method, we feed the factors, after some thresholding, to a schema that uses the ternary tessellation of Section \ref{sec:tess} and a parse-tree based permutation map (described in the supplement) to get sparse representations. We use these representations to extract exactly those items as would be extracted by the inverted index data structure applied to our sparse factors. The performance is compared against the following baselines- sign-random-projection hash (SRP-LSH) \cite{SRP-LSH}, Superbit-LSH \cite{SuperLSH}, Concomitant rank order statistics \cite{concom} and PCA-tree \cite{PCAtree}.

The comparison is two-fold. First, we compute the recovery accuracy- what proportion of relevant items was actually recovered by each of the above methods after discarding certain items. 
Next, for each of the methods, we compute for each user the proportion of items that are discarded. The percentage of items discarded showed a large variance for some of the baselines, hence we display them as histograms. 
The supplementary material also contains a set of figures that plot recovery accuracy against achieved sparsity, as well as the mean percentage of discarded items across methods for synthetic and real data.

Note that the percentage of items discarded has a direct relationship with the speed-up achieved- eg, if $\eta$ proportion of items are discarded, size of item list for score computation reduces to $(1 - \eta)$ which results in a $\frac{1}{1 - \eta}$-fold increase in speed.\vspace*{-0.2cm}

\subsection{Synthetic Data}\vspace*{-0.2cm}

For synthetic data, we randomly generate factors $\bU$ and $\bV$ using the standard normal distribution and construct the ``rating matrix" $\mathbf{R} = \bU\bV^\top$. We set the factor matrix $Z$ by concatenating the factors $\bU$ and $\bV$ as $\bZ = [\bU; \bV]$. Performance of different methods on these factors are evaluated with respect the true rating matrix $\mathbf{R}$. Experiments show that our methods achieves superior performance compared to the baselines, by obtaining both higher percentage of discarded items (figure \ref{fig:Simspar}) as well as higher accuracy (figure \ref{fig:SimAcc}). With close to 80\% of the items discarded on an average, our method achieves a nearly five-fold speed-up compared to the standard retrieval technique. \vspace*{-0.3cm}

\subsection{MovieLens Data}

We use the MovieLens100k dataset to learn low dimensional factors $\bU$ and $\bV$ for users and items respectively. The exact same procedure as for synthetic data is then repeated with the learned user and item factors.
Experiments show that for comparable performance in percentage of discarded items (figure \ref{fig:MLspar}), our method achieves much higher recovery accuracy (figure \ref{fig:MLAcc}) as compared to the baselines. With around 70\% items discarded on average, our method would result in over three-fold speed-up in retrieval.

\section{CONCLUSION AND FUTURE WORK}

In this manuscript we presented a novel framework that exploits structural properties of sparse vectors to significantly reduce the run time computational cost of factorisation models. We developed techniques that use geometry aware permutation maps on a tessellated unit sphere to obtain high dimensional sparse embeddings for factors with sparsity patterns related to angular closeness of the original factors. We also provided deterministic and efficient realisations for the framework. Future work for this would involve the design of better tessellation and sparse mapping schema for this framework and theoretical analyses of the same.

\subsubsection*{Acknowledgements}
This work was done while Avradeep Bhowmik and Nathan Liu were at Yahoo! Labs, Sunnyvale, CA.

\bibliographystyle{abbrv}
\bibliography{sparse}

\appendix

\section*{SUPPLEMENT}

\section{PROOFS}
\subsection*{Proof of Lemma 1}
\begin{proof}
Using standard Euclidean distance for projection of any factor $\bz$ on to the tessellating vectors $\Gamma$, recall by definition we have
\begin{eqnarray*}
\text{arg}\min_{\ba \in \Gamma} d(\ba, \bz) 
& = & \text{arg}\min_{\ba \in \Gamma} 1 - \frac{\ba^\top \bz}{\|\ba\|_2\|\bz\|_2} \\
& = & \text{arg}\max_{\ba \in \Gamma} \ba^\top \bz\ \ \ \ \ \ \ \ \ \ \ \left( \because \|\ba\|_2  = 1\right) \\
& = & \text{arg}\min_{\tilde{\ba} \in \mathcal{A}} 1 - \frac{\tilde{\ba}^\top \bz}{\|\tilde{\ba}\|_2\|\bz\|_2}
\end{eqnarray*}

Suppose $\ba$ has $t$ non-zero elements with the corresponding indices $I_\ba \subset \{1, 2, \cdots k\}$ with $| I_\ba | = t$. Clearly, the corresponding unnormalised $\tilde{\ba}$ would also have to have had $t$ non-zero elements, each of them $\pm 1$, and therefore $\|\tilde{\ba}\|_2 = \sqrt{t}$.

Therefore, we have $$\ba^\top \bz = \frac{\sum_{j \in I_\ba} sign(a^j) z^j}{\sqrt{t}}$$

Clearly, for any fixed $t$, the maximiser of the numerator is an $a$ such that each $a^j$ has the same sign as $z^j$ and $a^j$ is supported (non-zero) at the top $t$ elements (by absolute value) of $\bz$. Then, we have
\begin{eqnarray*}
\max_{\ba \in \Gamma} \ba^\top \bz &=& \max_{t} \max_{\ba: |I_\ba | = t} \frac{\sum_{j \in I_\ba} sign(a^j) z^j}{\sqrt{t}}\\
&=& \max_t\ z_s^t
\end{eqnarray*}
where $\bz_s$ is as defined in Algorithm \ref{alg:tess} of the main manuscript. This completes the proof of correctness of the projection operator.
\end{proof}
\subsection*{Proof of Lemma 2}

The steps to compute the approximately closest tessellation vector over $\Gamma_D$ are given in Algorithm \ref{alg:d-ary}. The proof given below is not the only one possible, other (possibly tighter) bounds can be obtained by using different proof techniques and different algebraic manipulations of the quantities involved.

\begin{proof}
Note that for any scalar $s$ with $|s| \leq 1$, there exists a scalar $h$ with $|h| \leq D$ such that $|s - \frac{h}{D}| < \frac{1}{D}$. Therefore, since each $\tba^j$ is a multiple of $\pm \frac{1}{D}$, for any $\bz \in \bS^k$, there exists $\tilde{\ba} \in \cA_D = \cB_D^k \setminus \{0\}^k$ such that $\|\bz - \tilde{\ba}\| = \sqrt{ \sum_i (\bz^i - \tba^i)^2} \leq \frac{\sqrt{k}}{D}$. 

For any vector $\bx \in \bS^k$, denote its projection on to $\cA_D$ as $$\cA_D(\bx) = \text{arg}\min_{\tilde{\ba} \in \cA_D} \|\tilde{\ba} - \bx \|_2$$
Clearly, by the preceding discussion,
\begin{equation}\label{eq:projD}
\|\bx - \cA_D(\bx)\|_2 \leq \frac{\sqrt{k}}{D}
\end{equation}

Moreover, $\cA_D(\bx)$ can be obtained by following steps \ref{alg:start} to \ref{alg:end} in \textsc{TessVector}-$D$($\bx, D$) as detailed in Algorithm \ref{alg:d-ary}.

Suppose for a factor $\bz$, the optimal projection on $\Gamma_D$ is
\begin{eqnarray*}
\ba_\bz^* &= \text{arg}\min_{\ba \in \Gamma_D} d(\ba, \bz) = \text{arg}\min_{\ba \in \Gamma_D} \frac{1}{2}\|\ba - \bz\|^2_2 \\ &= \text{arg}\min_{\ba \in \Gamma_D} \|\ba - \bz\|_2
\end{eqnarray*}

Suppose the projection obtained from \textsc{TessVector}-$D$($\bz, D$) is $\ba_\bz$. Then we have,
\begin{equation}\label{eq:proj2}
\|\bz - \ba^*_\bz\|_2 \leq \|\bz - \ba_\bz\|
\end{equation}

Now, we have
\begin{align}
\|\ba_\bz - \ba_\bz^*\| &\leq \|\ba_\bz - \bz\| + \|\bz - \ba_\bz^*\| && [\Delta\text{ ineq}] \\
  &\leq 2\|\ba_\bz - \bz\| && [\text{by \ref{eq:proj2}}] \\
  &\leq 2\left(\|\ba_\bz - \cA_D(\bz)\| + \|\cA_D(\bz) - \bz\|\right) && [\Delta\text{ ineq}]\label{eq:fin0}
\end{align}
Note that 
\begin{equation}
\ba_\bz = \frac{\cA_D(\bz)}{\|\cA_D(\bz)\|_2}
\end{equation}
Therefore, 
\begin{align*}
\|\ba_\bz - \cA_D(\bz)\| &= \|\frac{\cA_D(\bz)}{\|\cA_D(\bz)\|_2} - \cA_D(\bz)\| && \\
  &= \|\left(1 - \frac{1}{\|\cA_D(\bz)\|_2}\right)\cA_D(\bz)\| && \\
  &= |\|\cA_D(\bz)\|_2 - 1| && 
\end{align*}

Furthermore, by triangle inequality $$\|\cA_D(\bz)\|_2 \leq \|\cA_D(\bz) - \bz\|_2 + \|\bz\| = \|\cA_D(\bz) - \bz\|_2 + 1$$

Also, by triangle inequality, $$\|\cA_D(\bz)\|_2 \geq \| \bz \| - \|\bz  - \cA_D(\bz)\|_2 = 1 - \|\cA_D(\bz) - \bz\|_2$$

Therefore, $$ - \|\cA_D(\bz) - \bz\|_2 \leq \|\cA_D(\bz)\|_2 - 1 \leq \|\cA_D(\bz) - \bz\|_2$$
Hence,
\begin{equation}\label{eq:fin1}
|\|\cA_D(\bz)\|_2 - 1| \leq \|\cA_D(\bz) - \bz\|_2
\end{equation} 

Finally, by combining equations \ref{eq:fin0} and \ref{eq:fin1} with equation \ref{eq:projD}, we get that $\|\ba_\bz - \ba_\bz^*\|_2 \sim O(\frac{\sqrt{k}}{D})$. 

Since $d(\ba_\bz, \ba_\bz^*) \propto \|\ba_\bz - \ba_\bz^*\|^2_2$, we have our result $$d(\ba_\bz, \ba_\bz^*) \sim O\left(\frac{k}{D^2}\right)$$ Moreover, the time complexity of Algorithm \ref{alg:d-ary} is $O(k)$ and each step from \ref{alg:start1} to \ref{alg:end1} can be computed in parallel for each $j$ and each $\bz$. Also, algorithm \ref{alg:d-ary} requires no explicit storage of the tessellating set $\Gamma_D$. This completes the proof.
\end{proof}

\begin{algorithm}
\begin{algorithmic}[1]
\Procedure{TessVector-$D$}{$\bz, D$}
\State initialise $\tilde{\ba}_\bz$ to all zeros \label{alg:start}
\For{each $\bz \in \bZ$}
\For{each $j \in \{1, 2, 3, \cdots k\}$}\label{alg:start1}
\State compute $a_+ = |Dz^j - \ceil{Dz^j} |$ 
\State compute $a_- = |Dz^j - \lfloor Dz^j \rfloor |$ 
\If{$a_+ \leq a_-$}
\State set $\tilde{\ba}_z^j = \frac{\ceil{Dz^j}}{D}$ 
\Else
\State set $\tilde{\ba}_z^j = \frac{\floor{Dz^j}}{D}$
\EndIf
\EndFor\label{alg:end1}
\EndFor\label{alg:end}
\State normalise to get $\ba_\bz = \frac{\tilde{\ba}_\bz}{\|\tilde{\ba}_\bz \|_2}$
\State \textbf{return} $\ba_\bz$
\EndProcedure
\end{algorithmic}
\caption{Region Specification on $\Gamma_D$}
\label{alg:d-ary}
\end{algorithm}

\section{FURTHER DISCUSSION}

\subsection{Uniform Tessellation}
A key consideration while designing a tessellation schema on the unit sphere is whether the tessellation needs to be uniform or non-uniform over the surface of the unit sphere. There is no one way to capture the notion of ``uniformity" in the context of a tessellation schema. A few example conditions could be that each tessellating vector should be equidistant from the closest tessellating vector, or by symmetry should have the same number of closest tessellating vectors, or the diameter of each tile (distance between farthest points within the same tile) should be the same for each tile.

But whichever way ``uniformity" is defined, as a general scheme, a uniform tessellation would make intuitive sense because it captures the relevant locality properties of any set of factors irrespective of their distribution. However, in many instances a uniform tessellation may be overkill, and especially for clustered data, a non-uniform tessellation might be more appropriate from efficiency considerations. In particular, a uniform tessellation could be made into a non-uniform tessellation simply by dropping some of the tessellating vectors.

The directional tessellating set $\mathcal{A}$ on a ternary base set $\cB$ does not uniformly tessellate the unit sphere. This is because for each tessellating vector in $\Gamma$, the distance from the nearest tessellating vector depends on the number of non-zeros in the vector. 

In particular, the nearest neighbour to a vector $\ba_i$ is every vector $\ba_j$ such that the unnormalised vectors $\tilde{\ba}_i$ and $\tba_j$ differ by a Euclidian distance of 1 in the unnormalised space. That is, every nearest neighbour to a vector $\ba_i$ can be found by replacing a single element in the unnormalised version of the vector $\tba_i$ in the following way. First, obtain $\tba_j$ by replacing a single 1 or -1 in $\tba_i$ by a 0, or replace a single 0 by either a 1 or a -1. This is then re-normalised to get the corresponding $\ba_j$.

The proof of the above statement is the following. First, clearly the nearest neighbour to every $\ba_i$ must belong to the same orthant as $\ba_i$. Suppose $\ba_j$ belongs to the set of nearest neighbours. Therefore, $sign(\ba_i^\iota)sign(\ba_j^\iota) \geq 0$ for every $\iota = 1, 2, \cdots k$. Therefore, without loss of generality, assume that $\ba_i$ lies in the non-negative orthant, $\ba_j^\iota \geq 0 \ \forall \ \iota$.

Suppose $\ba_i$ has $t$ non-zero elements, and $\ba_j$ has $t+s$ non-zero elements, with $t > 0, s \neq 0$. Then, the angular distance between $\ba_i$ and $\ba_j$ is
\begin{equation*}
d(\ba_i, \ba_j) = 
\begin{cases}
1 - \sqrt{\frac{t}{t+s}} \ \ \text{ if } s > 0\\
1 - \sqrt{\frac{t-s}{t}} \ \ \text{ if } s < 0
\end{cases}
\end{equation*}

Clearly, the minimum distance is attained for any $t$ by setting $s = 1$. 

Following this, we see that the distance between closest neighbours $\ba_i, \ba_j$ is $d(\ba_i, \ba_j) = 1 - \sqrt{\frac{t}{t+1}}$. Therefore, the distance between closest neighbours for a tessellating vector depends on the number of non-zero elements in the vector. In particular, the set $\mathcal{A}$ is more densely packed with vectors oriented towards the ``centre" of each orthant as opposed to vectors along the axes or along any lower dimensional subspaces formed from subsets of the axes.

As mentioned in the main manuscript, obtaining uniform tessellations deterministically is a challenging task and heuristics \cite{tess0,tess1,tessicosa} must be resorted to.

\begin{figure*}[t]
\centering
\subfloat[subfig1][average percentage of discarded items for Synthetic Data]{
\includegraphics[width=0.5\textwidth, height = 5.5cm]{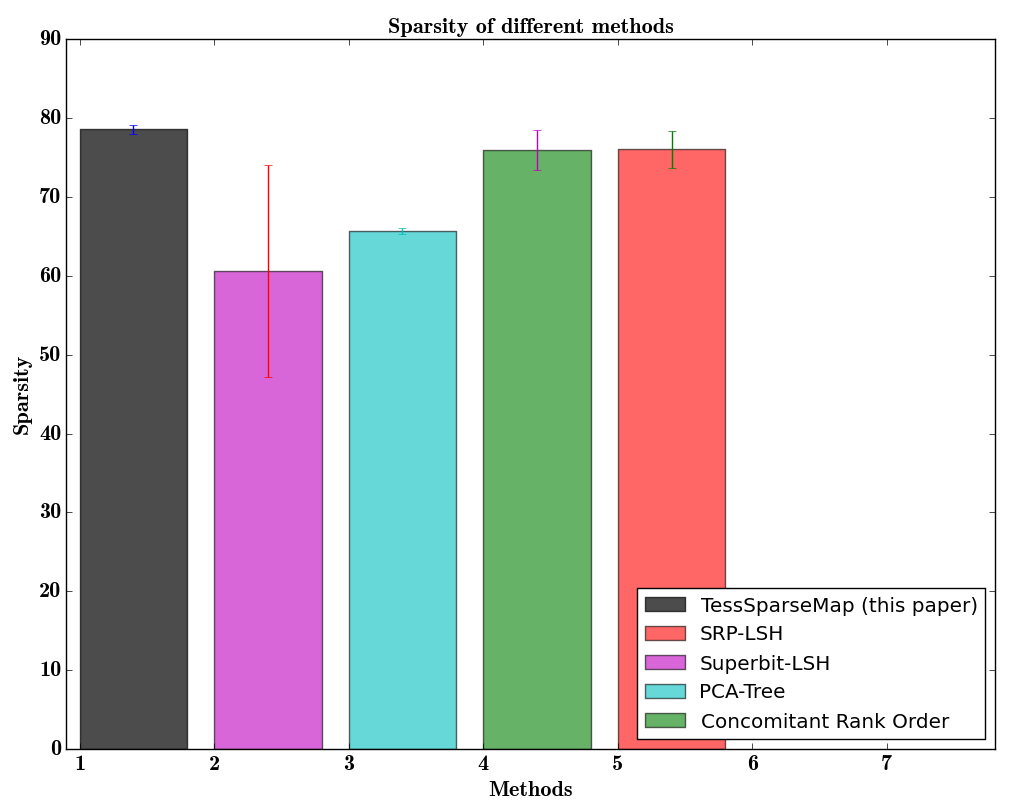}
\label{fig:Simbar}}
\subfloat[subfig2][average percentage of discarded items for MovieLens]{
\includegraphics[width=0.5\textwidth, height = 5.5cm]{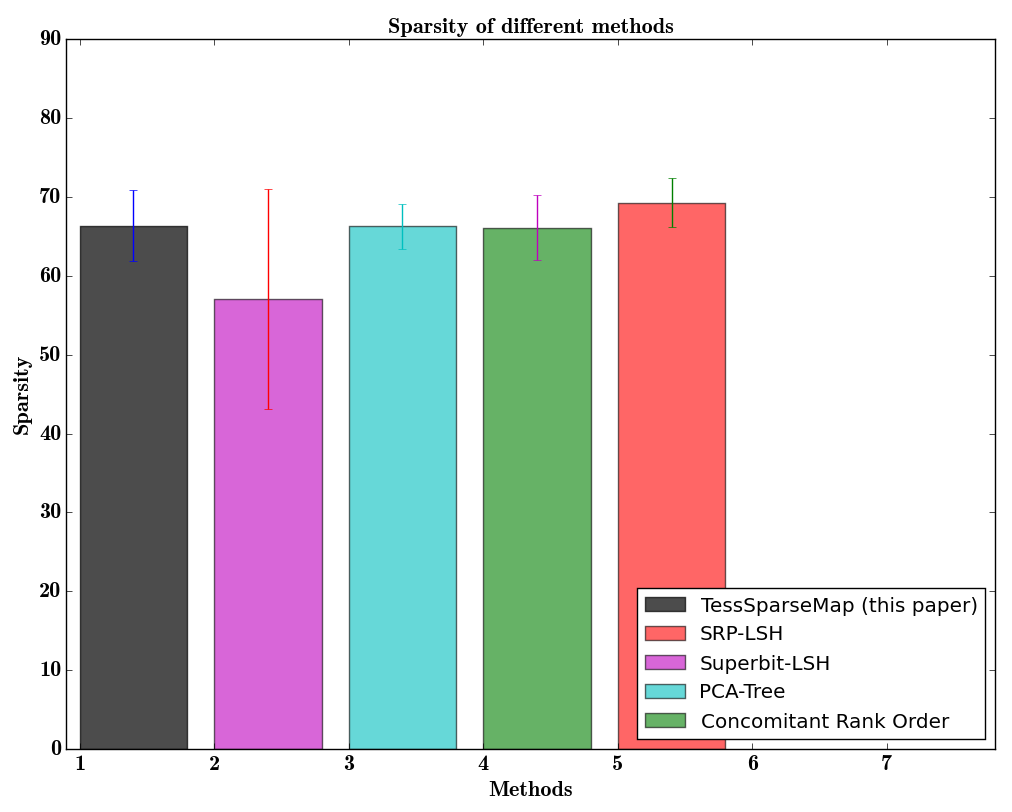}
\label{fig:MLbar}}
\caption{Mean percentage of discarded items across users for (a) Synthetic Data (b) MovieLens Data}
\label{fig:ResSpar}
\vspace*{-0.5cm}
\end{figure*}

%


\begin{figure*}[!htb]
\centering
\subfloat[subfig1][Recovery Accuracy]{
\includegraphics[width=0.5\textwidth, height = 5.9cm]{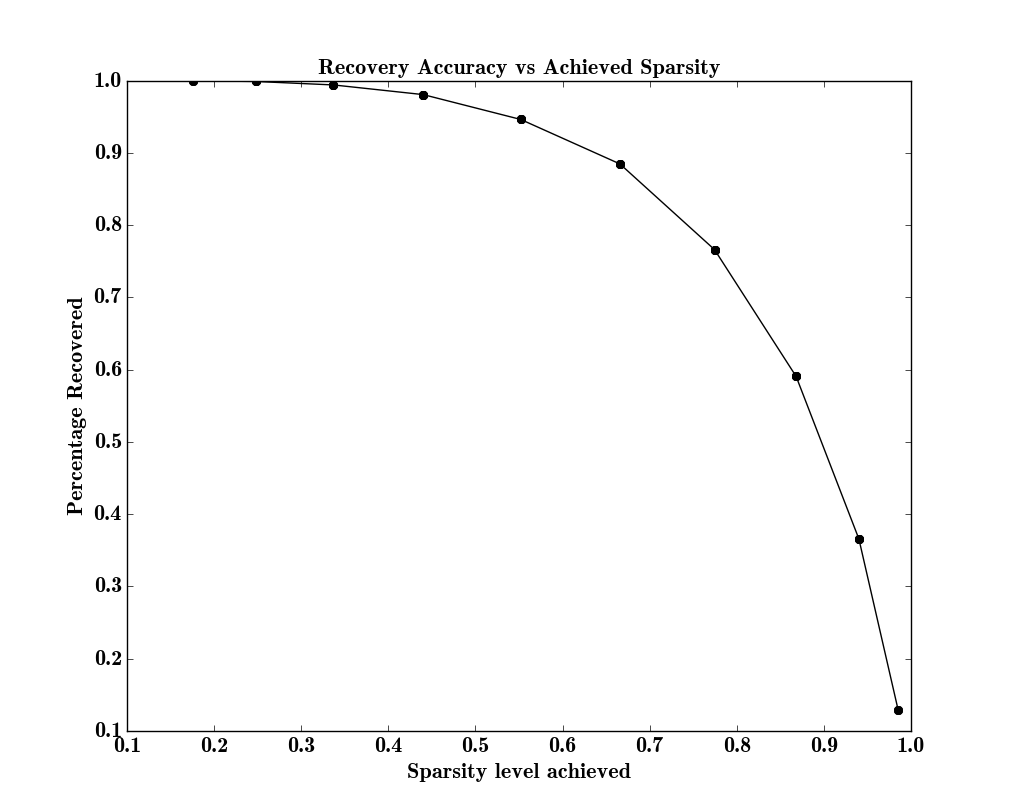}
\label{fig:SimRC}}
\subfloat[subfig2][Recovery Accuracy versus Sparsity]{
\includegraphics[width=0.5\textwidth, height = 5.9cm]{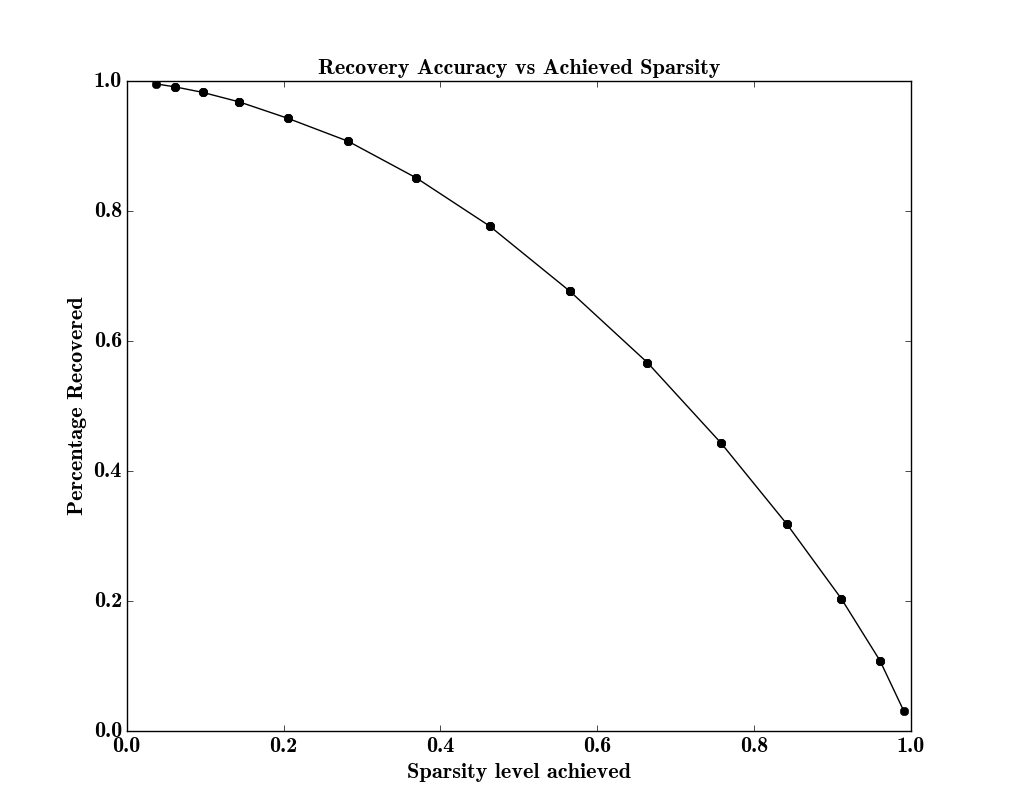}
\label{fig:MLRC}}
\caption{Plot for Recovery Accuracy versus Sparsity Achieved (a) Synthetic Data (b) MovieLens Data}
\label{fig:ResRC}
\vspace*{-0.5cm}
\end{figure*}

\subsection{Parse Tree Constructions}

In this section we describe some examples of parse-tree constructions for the permutation mapping step, in particular the parse tree used in the experiments in the main manuscript. Note that computing the permutation map proceeds in two steps- (i) reading $\ba_\bz$ at time $j$ as a sequence of $\delta$ characters at a time as $\tilde{\ba}^j_\delta = [\tba^{j-\delta}, \cdots \tba^j]$, and (ii) marking the next non-zero index via a counter $\tau_t$ on $\bphi(\cdot)$ as a function of $\tau_{j-1}$ and $\tba^\delta_j$ as $\tau_j = f(\tau_{j-1}; \tba_\delta^j)$. 

A key desideratum for our mapping scheme is that for any two $\ba, \ba'$ at any step $\tau_j = \tau_j'$ if and only if $[\ba^{j-t}, \cdots \ba^j] = [{\ba'}^{j-t}, \cdots {\ba'}^j]$ for some $t_0 \geq \delta$. This is useful in preventing ``accidental" overlapping sparsity, so that the same sparsity pattern is not obtained accidentally via two entirely different set of sliding window characters read on $\ba$ and $\ba'$.

It is immediately clear that the one-hot encoding satisfies this property with $t_0 = \delta = 1$. Another simple scheme (and one that we used in our experiments) is the following.

Consider a sliding window of size $\delta = 1$. Suppose after $j-1$ steps, the counter is at position $\tau_{j-1}$. Shift the counter to position $\tau_j$ depending on the value of currently read $\ba^j$ as follows-

\[ \tau_j =
  \begin{cases}
    kj       & \quad \text{if } \ba^j = 1\\
    \tau_{j-1} + 1  & \quad \text{if } \ba^j = 0\\
    k(k + j)  & \quad \text{if } \ba^j = -1\\
  \end{cases}
\]

The dimensionality increase required is $p \sim O(k^2)$, however, with the inverted index representation, we only require $O(klogk)$ storage space complexity.

Many other parse-tree methods are possible. In particular, a straightforward generalisation of the one-hot scheme described in the manuscript would obtain a class of methods that involve a one-hot encoding on a $D$-ary tessellation with a $\delta$-parse-tree which has $D^\delta$ leaf nodes. For this schema, for any two $\ba, \ba'$ we shall have the corresponding counters $\tau_j$ and $\tau'_l$ at time $j$ and l respectively to be equal $\tau_j = \tau'_l$ if and only if $j = l$ and ${\tba}^\delta_j = {\tba^{'\delta}_l}$.

\section{ADDITIONAL PLOTS}

We show some additional plots to augment the experimental results given in the main manuscript. Figure \ref{fig:Simbar} and \ref{fig:MLbar} respectively show the average sparsity levels obtained across all users for different methods mentioned in this mansucript. We also show error bars to give an idea of the variance.

Figure \ref{fig:SimRC} and \ref{fig:MLRC} show a plot of recovery accuracy plotted against average sparsity achieved across all users for our method.

\end{document}